\documentclass[10pt,twocolumn,letterpaper]{article}

\usepackage{iccv}
\usepackage{times}
\usepackage{stfloats}
\usepackage{graphicx}
\usepackage{amsmath}
\usepackage{amssymb}
\usepackage{xcolor}

\usepackage[pagebackref=true,breaklinks=true,letterpaper=true,colorlinks,bookmarks=false]{hyperref}

\newcommand{\modea}{\textbf{Mode A}}
\newcommand{\modeb}{\textbf{Mode B}}
\iccvfinalcopy 

\def\iccvPaperID{11291} 

\begin{document}

\title{RaViTT: Random Vision Transformer Tokens}

\author{Felipe A. Quezada\\
Institute of Biomedical Sciences,\\
Universidad de Chile\\
Santiago, Chile\\
{\tt\small felipe.quezada01@outlook.com}
\and
Carlos F. Navarro\\
Institute of Biomedical Sciences,\\
Universidad de Chile\\
Santiago, Chile\\
{\tt\small canavarr@ing.uchile.cl}
\and
Cristian Mu\~noz\\
Institute of Biomedical Sciences,\\
Universidad de Chile\\
Santiago, Chile\\
{\tt\small crmunoz.bustamante@gmail.com}
\and
Manuel Zamorano\\
Dept. of Electrical Engineering,\\
Universidad de Chile\\
Santiago, Chile\\
{\tt\small manuel.zamorano@ug.uchile.cl}
\and
Jorge Jara-Wilde\\
Institute of Biomedical Sciences,\\
Universidad de Chile\\
Santiago, Chile\\
{\tt\small mjjaraw@uchile.cl}
\and
Violeta Chang\\
Dept. of Informatics Engineering,\\
Universidad de Santiago\\
Santiago, Chile\\
{\tt\small violeta.chang@usach.cl}
\and
Crist\'obal A. Navarro\\
Institute of Informatics\\
Universidad Austral de Chile\\
Valdivia, Chile\\
{\tt\small cnavarro@inf.uach.cl}
\and
Mauricio Cerda\\
Institute of Biomedical Sciences,\\
Universidad de Chile\\
Santiago, Chile\\
{\tt\small mauricio.cerda@uchile.cl}
}

\maketitle

\begin{abstract}
Vision Transformers (ViTs) have successfully been applied to image classification problems where large annotated datasets are available. On the other hand, when fewer annotations are available, such as in biomedical applications, image augmentation techniques like introducing image variations or combinations have been proposed. However, regarding ViT patch sampling, less has been explored outside grid-based strategies. In this work, we propose Random Vision Transformer Tokens (RaViTT), a random patch sampling strategy that can be incorporated into existing ViTs. We experimentally evaluated RaViTT for image classification, comparing it with a baseline ViT and state-of-the-art (SOTA) augmentation techniques in 4 datasets, including ImageNet-1k and CIFAR-100. Results show that RaViTT increases the accuracy of the baseline in all datasets and outperforms the SOTA augmentation techniques in 3 out of 4 datasets by a significant margin ($+1.23$\% to $+4.32$\%). Interestingly, RaViTT accuracy improvements can be achieved even with fewer tokens, thus reducing the computational load of any ViT model for a given accuracy value.

\end{abstract}

\section{Introduction}

Deep Learning (DL) for image processing has progressed dramatically, yet unevenly, in the last decade. For instance, DL tools for biomedical applications, such as digital pathology, have shown a relatively slow adoption \cite{Campanella2019}. One of the critical reasons for this is the severe lack of large annotated datasets compared with other domains \cite{Campanella2019}: digital pathology requires extensive cell- or tile-level expert annotations in large Whole Slide Images (WSIs) of about $100,000 \times 100,000$ pixels, to train supervised models. This is an example of a non-trivial classification scenario where it becomes very challenging to compile large datasets \cite{Kanavati2020}.

One approach to artificially increase the size of a dataset is to generate image variations from the samples based on transformations such as flipping, rotation, and scaling.
These variations make the models more robust to the types of variations generated, improving the model's generalization performance and 
reducing overfit. More recently, these approaches have been integrated into pipelines with parameters subject to optimization, such as RandAugment \cite{randaugment}. 
Also, the creation of additional samples by merging images and labels from the original dataset has been proposed \cite{Shorten2019}.

Interestingly, before the broad adoption of DL, the extraction of randomized subwindows has been proposed as a technique to artificially increase the training sample size \cite{desir2012classification,navarro2019color}. Randomized subwindows are defined as small to contain one texture/object, and due to randomness, avoid the definition of rigid grids to define image patches. However, how to integrate patch randomization ideas into state-of-the-art image classification methods has not been addressed.

In image classification tasks, the state of the art (SOTA) comes from Vision Transformer (ViT) models, inspired by natural language processing to contextualize long texts, adapted to contextualize image patches \cite{dosovitskiy2020image}. Despite their recent introduction, ViT models have already been explored in multiple patching strategies such as selection~\cite{meng2022adavit}, multi-scale~\cite{ren2022beyond}, or hierarchical representation~\cite{chen2022scaling}. Yet, these methods keep using regular grid patterns to generate image patch tokens.

\begin{figure}[t]
\begin{center}
\includegraphics[width=\linewidth]{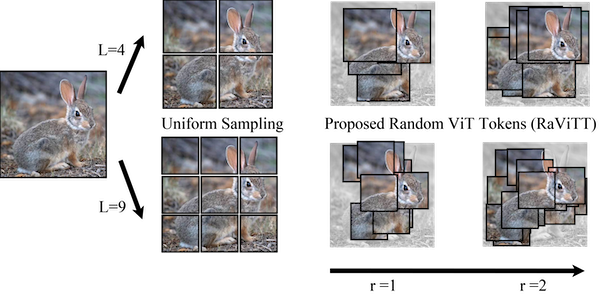}
\end{center}
   \caption{Uniform sampling and the proposed token sampling. $L$ is the number of tokens per image, $r$ is the sampling factor.} 
\label{fig:general}
\end{figure}

In this work, we propose a random patch sampling strategy for ViT models as a general way to improve image classification performance (Figure \ref{fig:general}). The strategy can use any ViT as backbone. Our main results include the accuracy increase in all evaluated datasets with respect to the baseline, outperforming SOTA augmentation techniques in 3 out of 4 datasets ($+1.23$\%, $+3.66$\%, $+4.32$\%). Moreover,  accuracy improvements can be achieved with fewer tokens than the baseline.

We organize the remainder of this article as follows: Section 2 reviews existing augmentation methods and related work. Section 3 describes ViTs, regular uniform and the proposed patch sampling strategy. Sections 4 and 5 describe our evaluation methodology and main results, respectively. Section 6 presents insights into how randomization relates to ViT models. Conclusions are presented in Section 7.

\section{Related works}
This section reviews existing data augmentation techniques using single or multiple images and related patch handling methods.

\subsection{Single image augmentation}
Standard data augmentation techniques for image processing may involve flipping, rotating, scaling, cropping, translating, adjusting the color, adding noise, Gaussian blurring, shearing, and more. Although such approaches have been shown to increase the accuracy of Long Short-Term Memory (LSTM) based architectures by up to 11.5\% on the CIFAR-10 dataset \cite{SimpleDataAug}, there are no precise criteria for determining specific parameters that are required by these methods. Due to this lack of a recipe, recent architectures rely on automated selection methods, such as RandAugment \cite{randaugment}.

RandAugment is an augmentation technique for image data that automatically selects a set of image transformations to be applied randomly and variably during training, using a predefined set of operations and hyper-parameters that control the strength and number of transformations. The selection leads to a diverse and representative training set that improves the performance of DL models on various computer vision tasks, including image classification, object detection, and semantic segmentation. For example, the authors of RandAugment reported a significant 1.5\% improvement in accuracy on CIFAR-10 with a ResNet-50, while on ImageNet-1k, it improved the Top-1 accuracy of an EfficientNet-B7 model by 1.0\%. However, the improvement magnitude strongly depends on the architecture and dataset.

\subsection{Multiple image augmentation}
Recent works on data augmentation that have gained popularity focus on generating new images: MixUp \cite{zhang2017mixup} combines two data points by interpolating their features, and CutMix \cite{yun2019cutmix} creates new images by cutting and pasting regions of existing images. Both methods merge labels and images with different approaches, increasing the classification accuracy for ImageNet-1k by 1.1\% and 2.3\%, respectively, utilizing a ResNet-50 architecture. Generative Adversarial Networks (GANs) have been applied to generate new images that can be used to augment the training data \cite{Shorten2019}, decreasing the error in 1.6\% for CIFAR-10 and 3.9\% for CIFAR-100. Lastly, among methods exclusively designed for ViT, TokenMix \cite{liu2022tokenmix} performs data augmentation by mixing two images blending their feature maps, which are obtained by feeding the images through a ViT model. TokenMix applied to the DeiT-T model \cite{deit2021} showed an increase of 0.5\% in accuracy from previous results obtained with CutMix.

\subsection{Patch handling methods in ViTs} %

 


ViT models are computationally intensive \cite{dosovitskiy2020image} due to the nonlinear increase in load as the number of image patches increases. To improve ViT inference efficiency, an adaptive computational framework was proposed \cite{meng2022adavit}, which learns to derive usage policies on patches, transformer blocks, and self-attention heads, with minimal drops in accuracy for image recognition.


Dynamic Window Visual Transformer (DW-ViT) \cite{ren2022beyond} looks at the impact of window size on model performance. It obtains multi-scale information by assigning windows of different sizes to different head groups of window multi-head self-attention. The window size significantly influences the model's performance. DW-ViT achieves a significant improvement of $3.3$\% Top-1 Accuracy compared to methods of single-scale windows \cite{ren2022beyond}.


Because of their considerable computational costs, ViTs have been generally researched employing low-resolution images. For large images such as WSIs, specific architectures were introduced: HIPT \cite{chen2022scaling} is a ViT that harnesses the hierarchical WSI structure, utilizing two levels of self-supervised learning to learn high-resolution image representations. HIPT achieves a performance increase of 1.86\%, 2.59\%, 0.72\% on three carcinoma databases respectively using 100\% of training data. When the training data is reduced to 25\% as 10-fold cross-validated AUC, furthers increases of 3.14\%, 8.33\%, 1.78\% are achieved, respectively \cite{chen2022scaling}. Liu \etal \cite{liu2021iccv} introduced a hierarchical ViT with layers of shifted, non-overlapping windows for computing self-attention. Windows across layers are connected by extending beyond the boundaries of the previous windows.




\begin{figure*}[t]
  \centering
  \includegraphics[scale=0.37]{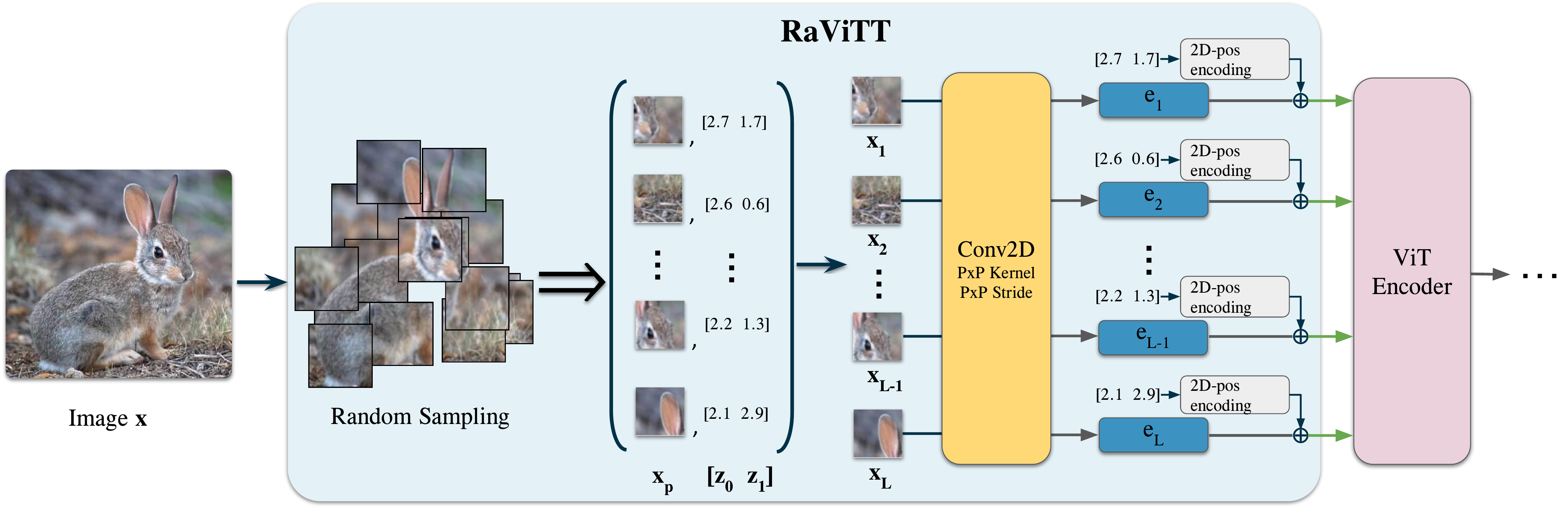}
  \caption{Summary of the proposed RaViTT method.}
  \label{fig:RaViTT}
\end{figure*}

\section{Proposing Random ViT Tokens (RaViTT)}

This section introduces regular uniform ViT sampling and details our proposed Random ViT Token method.

\subsection{Regular uniform ViT sampling}
ViT models typically rely on a regular uniform sampling pattern when extracting patches as tokens from an input image \cite{dosovitskiy2020image}. As a result, sampling in a regular grid-like manner ensures that the model can \textit{see} the entire image while maintaining a relatively low sequence length.
Let $P$ the patch resolution size $(P, P)$, and $x$ an image $x \in \mathbb{R}^{C \times H \times W}$, where $C$, $H$ and $W$ are the channel number, height, and width, respectively, regular uniform sampling works as follows: a regular uniform patch sampling technique consists of slicing the input image $x$ into a sequence of $L=\frac{HW}{P^2}$ non-overlapping individual patches $x_1, x_2, \dots, x_p, \dots, x_L$ where $x_p \in \mathbb{R}^{(P^2 \cdot C)}$.


The image patches are extracted and listed to form the input token sequence of the ViT model, with size $L$. The ViT processes the input, typically through either a linear or convolutional layer, creating an abstract representation of each patch, called the embedding vector $e_p$. To preserve 2D coherence, a positional encoding vector specifically designed to be aware of the 2D nature of the input data is added to each embedding vector $e_p$. As a result, each embedding vector has a position assigned to it. An alternative option is to include a trainable position encoding parameter for each embedding vector to enable automatic modeling of the 2D data \cite{dosovitskiy2020image}. After creating the embedding vectors $e_p$ and encoding their position, the transformer divides them into its $h$ heads and proceeds with the standard pipeline.


\subsection{Proposed RaViTT method}
Current patch sampling in ViTs relies on regular grid-like sampling techniques. Despite previous research showing positive results from random sampling \cite{navarro2019color}, these approaches have not been explored in ViTs. 

Random ViT Tokens (RaViTT) is an alternative patch sampling strategy that employs random sampling instead of the standard regular uniform sampling approach typically used by ViTs. 
By randomly sampling patches, it is possible to replicate and to center relevant features or phenomena within individual patches. Additionally, this approach can expand the space of possible inputs for the linear projection model, as it allows for a broader range of patch locations to be considered. This increased flexibility could improve the model's ability to extract essential features from the input data, leading to better performance on downstream tasks. The approach could be especially beneficial for datasets with limited training samples or objects at the patch level.


The core idea behind RaViTT (Figure \ref{fig:RaViTT}) is to randomly select $r\cdot L$ patches of $(P,P)$ pixels from the input image $x$, allowing overlapping among patches to occur. The hyper-parameter $r$ controls the number of patches sampled. Each patch $x_p$ is selected from the input image $x$ with random coordinates $(z_0, z_1) \in \mathbb{R}^2$ in patch space $[0,\sqrt{L}-1] \times [0,\sqrt{L}-1]$. That is, $z_0 \sim \mathcal{U} (0, \sqrt{L}-1)$ and $ z_1 \sim \mathcal{U}(0, \sqrt{L}-1)$. 
As a note, these patch coordinates refer to the top-left corner of the patch in pixel space, which is $(z_0P, z_1P)$. 
A significant distinction between RaViTT and uniform sampling is that RaViTT does not limit the number of patches that can be taken from the input image given its size $(H \times W)$ and $P$. That is, RaViTT is not restricted to selecting $L$ patches,
since $r\cdot L$ can be as low as $1$ and as large as needed. 

Sampling patches from random positions on the input image $x$ introduces an inconsistency with regular 2D positional encoding, as the order of patches does not necessarily correlate with their position on $x$. Typical 2D position encoders assume a regular $\sqrt{L}\times \sqrt{L}$ grid of patches, which may not be the case for some values of $r$. 
The patch position of each patch $x_p$ at the time of sampling is stored to avoid assuming a regular grid.
Position information is used to compute the 2D $\sin$/$\cos$ positional encoding of each patch separately. In doing so, the model ensures that the 2D spatial nature of the data is preserved during processing.

At implementation level, the RaViTT random sampling can be done efficiently using the PyTorch \texttt{grid\_sample} function (or similar), with a random list of $z_0$ and $z_1$ positions as input. 
It is noteworthy that in RaViTT the random sampling yields overlapped patches in most cases (unless $r$ is significantly low). 


The RaViTT approach can be combined with any variation of ViT as backbone model, with the potential to be used in a wide range of applications.
RaViTT can be used in two modes: \modea{} for both training and testing phases; and \modeb{} only for training, with the testing phase using the regular uniform sampling ViT backbone.


\section{Experimental design}

RaViTT performance was evaluated in terms of Top-1 accuracy (\%), in comparison with the original ViT \cite{dosovitskiy2020image} which is a known reference. Comparisons were made upon $4$ datasets (Section \ref{sec-datasets}), with images that vary from $32\times32$ pixels to $500\times500$ pixels. Additionally, results with and without SOTA data augmentation methods (Section \ref{sec-trainingdetails}) are presented.


\subsection{Datasets}
\label{sec-datasets}
Four datasets of color images were used for testing.

\begin{itemize}

    \item \textbf{ImageNet-1k:}  large-scale image database that contains approximately 1.28 million training images and 50K validation images of variable patch size (average $469\times387$ pixels) across 1000 classes; it has been widely used for developing and testing image classification as well as pre-training models \cite{ImageNet_dataset}. For training and evaluation, we used the fixed ``train'' and ``validation'' splits, respectively.

    \item \textbf{CIFAR-100:} $60,000$ small images ($32\times32$ pixels) from 100 classes, divided in two subsets: $50,000$ training images and $10,000$ validation images. CIFAR-100 is commonly used for training and evaluating computer vision models for fine-grained classification tasks \cite{Cifar_dataset}. For training and evaluation, we used the fixed ``train'' and ``test'' splits, respectively.

    \item \textbf{G. CANCER:} $34$ WSIs of clinical Gastric CANCER resections \cite{ecancer} (Clinical Trial NCT0163320). After the acquisition, experts labeled small areas, which were divided into $1723$ image patches ($299\times299$ pixels) with $3$ possible categories with high imbalance ($86$\%, $6$\%, $8$\%) (G. CANCER-3). For training and evaluation, we performed a 80-20 split, respectively. Every model was trained with the same images.

    \item \textbf{DeFungi:} $660$ fungal images divided in $9114$ patches of size $500\times500$ pixels, with $5$ different fungal species class labels, used to train and evaluate computer vision models for fungi identification \cite{DeFungi_dataset}. For training and evaluation, we performed a 85-15 split, respectively. Every model was trained with the same images.
\end{itemize}

\begin{figure}[ht!]
\begin{center}
\includegraphics[width=\linewidth]{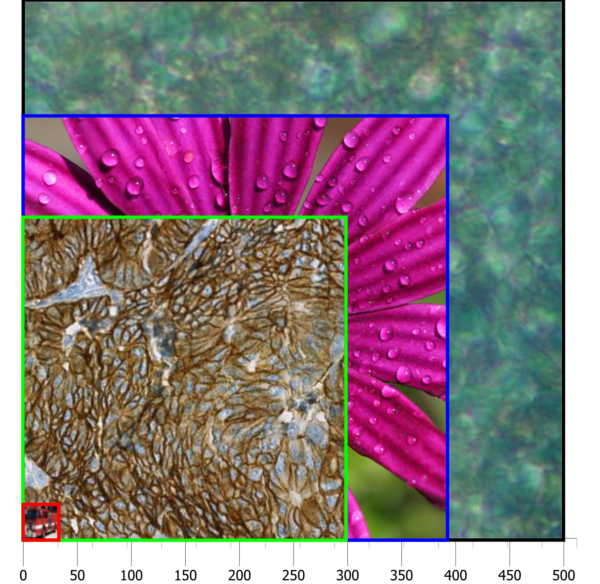}
\end{center}
   \caption{Example image patches from the datasets used: CIFAR-100 (red square), $32\times32$ pixels size; G. CANCER (green square), $299\times299$ pixels; ImageNet-1k (blue square), $469\times387$ pixels on average; and DeFungi (dark purple square), $500\times500$ pixels.}
\label{fig:datasets-scales}
\end{figure}

Figure \ref{fig:datasets-scales} illustrates the difference in image scale for the datasets. The patch size of CIFAR-100 (red square) is tiny compared to the others.

\subsection{Training details}
\label{sec-trainingdetails}
The training scheme follows many of the recommended design guidelines of Beyer \etal \cite{vit_baseline} to achieve better performance.
\begin{itemize}
    \item \textbf{Architecture:} testing uses the ViT-S/16 model as backbone (23M params $\sim 83\text{MB}$), as it has been shown to achieve good results in a reasonable amount of time \cite{vit_baseline}. We used a convolutional layer as patch tokenizer/embedder and 2D $\sin$-$\cos$ positional encoding instead of trainable parameters. For classification, we used the global average pool (GAP) technique and a single linear layer with an equal number of outputs as the number of classes.
    
    \item \textbf{Data:} both training and validation images were re-scaled to $224\times 224$ pixels and normalized in the range $[-1,1]$. Due to the backbone and image size used, the base sequence length was $L=14\times 14=196$. RaViTT was tested with values of $r \in \{0.5, 1, 2, 3, 4\}$, rendering the input length in it as $r\cdot L \in \{98, 196, 392, 588, 784\}$. 
    
    \item \textbf{Hyper-parameters:} we used the Adam optimizer with a weight decay of $\lambda = 0.0001$, and $\beta_1 = 0.9, \beta_2=0.999$. Due to hardware limitations, the batch size was reduced from $1024$ to $512$, and the learning rate $\alpha$ was readjusted from $0.001$ to $0.0007$ following \cite{granziol2022learning}. 
    The training was performed once without dropout for fixed $90$ epochs, employing cosine decay and a linear warm-up for the first $4.44\%$ of the total steps, as proposed by the guidelines \cite{vit_baseline}. The loss function used was cross-entropy.  

    
    \item \textbf{Data augmentation:} two schemes were considered; RandAugment $+$ Random Flip (referred to only as RandAugment) and MixUp. RandAugment was set with $n=2$ operations and magnitude $m=10$, and Random Flip was set with probability $p=0.5$. As for MixUp, it was set with parameter $\alpha=0.2$. 
\end{itemize}
RaViTT was evaluated using the two modes defined, \modea{} (random sampling for training and evaluation) and \modeb{} (random sampling for training, ViT backbone sampling for testing).
Next, the best version of RaViTT and the base ViT-S/16 were trained with RandAugment and MixUp.


\subsection{Software and hardware setup}
\textbf{Software:} source code written in Python 3.10.6 using PyTorch 1.13.1. PyTorch's \textit{Automatic Mixed Precision} feature was enabled to take advantage of GPU Tensor Cores. CUDA version 11.7 and the operating system Ubuntu Linux 20.04 64-bit were used.

\textbf{Hardware:} a NVIDIA DGX A100 system with $8\times$ A100 GPUs (each one with 6912 CUDA cores, 432 Tensor cores and 40 GB of HBM2e memory), $2\times$ AMD EPYC 7742 64-core CPUs, and 1 TB of system RAM. With this setup, training RaViTT for 90 epochs on ImageNet-1k took from $8$ hours ($r=0.5$) to $28$ hours ($r=4$), approximately. For the other datasets, running times were significantly lower (to the order of hours), even with $r=4$.


\section{Results}
Section \ref{sec-singledataset} with Tables 1-4 present the results for each dataset. Each table shows the model's Top-1 accuracy (\%) and the inference Giga Floating Point Operations (GFLOPs) for the baseline ViT-S/16 model and RaViTT, for $r\in\{0.5, 1, 2, 3, 4\}$, using \modea{} and \modeb{}. Section \ref{sec-overall} with Table \ref{table-overall} present an overall evaluation, comparing RaViTT with SOTA data augmentation methods.

\subsection{Single dataset results}
\label{sec-singledataset}
    \begin{table}[ht!]
    \begin{center}
    \resizebox{\columnwidth}{!}{

    \begin{tabular}{|l|c|c||c|c|}

    \hline
    Model & \multicolumn{2}{c||}{Top-1 Acc (\%)} & \multicolumn{2}{c|}{Inference GFLOPs}\\
    \hline\hline
    ViT-S/16 \cite{dosovitskiy2020image} & \multicolumn{2}{c||}{ 54.10 } & \multicolumn{2}{c|}{  4.6  }\\
    \hline    
    \hline

    RaViTT & \modea{} & \modeb{} & \modea{} &  \modeb{} \\
    \hline

    \multicolumn{1}{|l|}{$r=0.5$} & 47.48 & 51.66  & 2.2 & 4.6 \\
    \multicolumn{1}{|l|}{$r=1$} & 52.47 & \textbf{55.00}  & 4.6 & 4.6\\
    \multicolumn{1}{|l|}{$r=2$} & 57.55 & 53.95  & 9.9 & 4.6\\
    \multicolumn{1}{|l|}{$r=3$} & 59.43 & 48.41 & 15.9 & 4.6\\
    \multicolumn{1}{|l|}{$r=4$} & \textbf{59.86} & 42.58 & 22.6 & 4.6\\
    \hline
    \end{tabular}
    }
    \end{center}
    \caption{ImageNet-1k results.}
    \label{table-imagenet}
    
    \end{table}

Results obtained with ImageNet-1k are shown in Table \ref{table-imagenet}. With a base Top-1 accuracy (\%) of $54.1$\% in the regular ViT-S/16, RaViTT with \modea{} shows increasingly higher performance for $r\ge2$, with a peak of $59.86$\% for $r=4$, while RaViTT with \modeb{} was outperformed by the baseline ViT for all values of $r$, except for $r=1$ with $55$\%.

    \begin{table}[ht!]
    \begin{center}
    \resizebox{\columnwidth}{!}{

    \begin{tabular}{|l|c|c||c|c|}
    \hline
    Model & \multicolumn{2}{c||}{Top-1 Acc (\%)} & \multicolumn{2}{c|}{Inference GFLOPs}\\
    \hline\hline
    ViT-S/16 \cite{dosovitskiy2020image} & \multicolumn{2}{c||}{ 51.91 } & \multicolumn{2}{c|}{  4.6  }\\
    \hline\hline
    RaViTT & \modea{} &  \modeb{} & \modea{} &  \modeb{} \\
    \hline
    \multicolumn{1}{|l|}{$r=0.5$} & 57.40 & \textbf{60.16}  & 2.2 & 4.6\\
    \multicolumn{1}{|l|}{$r=1$}  & 58.69 & \textbf{60.16} & 4.6 & 4.6\\
    \multicolumn{1}{|l|}{$r=2$}  & 59.74 & 59.75 & 9.9 & 4.6\\
    \multicolumn{1}{|l|}{$r=3$}  & \textbf{59.59} & 58.57 & 15.9 & 4.6 \\
    \multicolumn{1}{|l|}{$r=4$}  & 59.19 & 57.76 & 22.6 & 4.6 \\
    \hline
    \end{tabular}
    }
    \end{center}
    \caption{CIFAR-100 results.}
    \label{table-cifar}
    \end{table}

Results with CIFAR-100 are presented in Table \ref{table-cifar}. Here, all results from RaViTT outperform the baseline ViT with a peak of $60.16$\% for $r=0.5$ and $r=1$ using \modeb{}. Nonetheless, the rest of the results are relatively close, within $3$ points distance. It is worth noting that switching from \modea{} to \modeb{} tends to increase performance for $r \le 2$, while the opposite is observed for $r > 2$.


    \begin{table}[ht!]
    \begin{center}
    \resizebox{\columnwidth}{!}{

    \begin{tabular}{|l|c|c||c|c|}
    \hline
    Model & \multicolumn{2}{c||}{Top-1 Acc (\%)} & \multicolumn{2}{c|}{Inference GFLOPs}\\
    \hline\hline
    ViT-S/16 \cite{dosovitskiy2020image} & \multicolumn{2}{c||}{  92.14  } & \multicolumn{2}{c|}{  4.6  }\\
    \hline\hline
    RaViTT & \modea{} &  \modeb{} & \modea{} &  \modeb{} \\
    \hline
    \multicolumn{1}{|l|}{$r=0.5$} & \textbf{93.61} & 93.36 & 2.2 & 4.6\\
    \multicolumn{1}{|l|}{$r=1$}   & 93.36 & \textbf{94.35} & 4.6 & 4.6 \\
    \multicolumn{1}{|l|}{$r=2$}   & \textbf{93.61} & 92.62 & 9.9 & 4.6 \\
    \multicolumn{1}{|l|}{$r=3$}   & \textbf{93.61} & 93.12 & 15.9 & 4.6 \\
    \multicolumn{1}{|l|}{$r=4$}   & 93.36 & 93.61 & 22.6 & 4.6 \\
    \hline
    \end{tabular}
    }
    \end{center}
    \caption{G. CANCER-3 results.}
    \label{table-cancer}
    \end{table}
Results with the G. CANCER-3 dataset are shown in Table \ref{table-cancer}. Accuracy values across all models are very similar, with slight differences. The baseline ViT-S/16 achieves the lowest accuracy, $92.14$\%. RaViTT achieves similar accuracy across $r$ using \modea{}, with the highest accuracy value $93.61$\% shared by $r={0.5, 2, 3}$. RaViTT with \modeb{} yields the highest accuracy; $94.35$\% at $r=1$.


    \begin{table}[ht!]
    \begin{center}
    \resizebox{\columnwidth}{!}{

    \begin{tabular}{|l|c|c||c|c|}
    \hline
    Model & \multicolumn{2}{c||}{Top-1 Acc (\%)} & \multicolumn{2}{c|}{Inference GFLOPs}\\
    \hline\hline
    ViT-S/16 \cite{dosovitskiy2020image} & \multicolumn{2}{c||}{ 76.9 } & \multicolumn{2}{c|}{  4.6  }\\
    \hline\hline
    RaViTT & \modea{} &  \modeb{} & \modea{} &  \modeb{} \\
    \hline
    \multicolumn{1}{|l|}{$r=0.5$} & 85.67 & \textbf{86.50}  & 2.2 & 4.6 \\
    \multicolumn{1}{|l|}{$r=1$}   & \textbf{86.78} & 86.04  & 4.6 & 4.6\\
    \multicolumn{1}{|l|}{$r=2$}   & 84.80 & 84.65  & 9.9 & 4.6 \\
    \multicolumn{1}{|l|}{$r=3$}   & 85.45 & 83.33 & 15.9 & 4.6 \\
    \multicolumn{1}{|l|}{$r=4$}   & 82.68 & 82.82 & 22.6 & 4.6  \\
    \hline
    \end{tabular}
    }
    \end{center}
    \caption{DeFungi results.}
    \label{table-defungi}
    \end{table}

Results with the DeFungi dataset are presented in Table \ref{table-defungi}. Here, RaViTT improves the performance of the baseline ViT-S/16, which achieved $76.9$\%. RaViTT with \modea{} for $r=1$ yields the best accuracy with $86.78$\%. However, as $r$ increases, the accuracy decreases to a minimum of $82.68$\% for $r=4$. RaViTT in \modeb{} shows a similar trend, with peak accuracy of $86.5$\% for $r=0.5$, which also decreases as $r$ increases, to a minimum of $82.82$\% for $r=4$.

Along Tables \ref{table-imagenet}, \ref{table-cifar}, \ref{table-cancer}, and \ref{table-defungi}, GFLOPs at inference time are reported. As it can be observed, operations increase linearly with $r$, with a slight overhead. Also, RaViTT with \modeb{} shows no inference overhead in terms of GFLOPs, which are the same for the backbone ($4.6$ GFLOPs).


\subsection{Overall evaluation}
\label{sec-overall}
As shown in Table~\ref{table-overall}, comparing the base ViT-S/16 model with RaViTT, accuracy is increased in the four evaluated datasets from $2.21$\% (G. CANCER-3) up to $9.88$\% (DeFungi). Compared with SOTA data augmentation methods, in $3$ of $4$ datasets RaViTT yields higher accuracy than RandAugment (ViT-S/16 + RandAugment) and MixUp (ViT-S/16 + RandAugment + MixUp). Specifically, measured increases on accuracy are $+3.66$\% in ImageNet-1k $\Delta$($60.52$\% - $56.20$\%), $+1.23$\% in G. CANCER-3 $\Delta$($94.35$\% - $93.12$\%), and $+4.32$\% in DeFungi $\Delta$($86.78$\% - $82.46$\%). CIFAR-100 is the only dataset where RaViTT shows a lower accuracy increase than the baseline augmentation. We note that the size of CIFAR-100 input images is small ($32\times32$ pixels, Figure~\ref{fig:datasets-scales}), and the ViT tokens approach pixel size ($4\times4$ pixels), where overlap becomes similar to repetition. 

 We also observed that the advantage of combining techniques varies among datasets. When RaViTT is combined with existing RandAugment and MixUp, in $2$ of the $4$ datasets we found benefits in combining methods (RaViTT + RandAugment, Table~\ref{table-overall}).

    \begin{table*}[ht!]
    \begin{center}
    \resizebox{\textwidth}{!}{
    \begin{tabular}{|c|c|c||c|c|}
    \hline
    Model \textbackslash \ Dataset & ImageNet-1k & CIFAR-100 & G. CANCER-3  & DeFungi \\ 
    \hline\hline
    \multicolumn{1}{|l|}{ViT-S/16 \cite{dosovitskiy2020image}}                    & 54.10 (+00.00) & 51.91 (+00.00)& 92.14 (+00.00)  & 76.90 (+00.00) \\
    \multicolumn{1}{|l|}{ViT-S/16 \cite{dosovitskiy2020image} + RandAugment\cite{randaugment}}          & 56.20 (+02.10) & 69.94 (+18.03)     & 93.12 (+00.98)       & 82.46 (+05.56) \\
    \multicolumn{1}{|l|}{ViT-S/16 \cite{dosovitskiy2020image} + RandAugment\cite{randaugment} + MixUp\cite{zhang2017mixup}}  & 55.23 (+01.13) & \textbf{71.00 (+19.09)}     & 92.38 (+00.24)       & 76.02  (-00.88) \\
    \hline
    Best RaViTT & ImageNet-1k [$r=4$] & CIFAR-100 [$r=1$] & G. CANCER-3 [$r=1$] & DeFungi [$r=1$]\\ 
    \hline
    \multicolumn{1}{|l|}{RaViTT} & 59.86 (+05.76) & 59.74 (+07.83)& \textbf{94.35 (+02.21)} & \textbf{86.78 (+09.88) }  \\
    \multicolumn{1}{|l|}{RaViTT + RandAugment \cite{randaugment}}  & \textbf{60.52 (+06.42)} & 70.42 (+18.51)& 92.87 (+00.73) & 79.53 (+02.63)  \\
    \multicolumn{1}{|l|}{RaViTT + RandAugment\cite{randaugment} + MixUp \cite{zhang2017mixup}} & 57.11  (+03.01)& 69.43 (+17.52)& 92.14 (+00.00) & 78.22  (+01.32) \\
    \hline
    \end{tabular}
    }
    \end{center}
    \caption{Top-1 Accuracy (\%) of models across datasets. Values within parentheses indicate the improvement relative to the baseline. When more than one model achieves the same performance, the one with higher RandAugment accuracy is selected.}
    \label{table-overall}
    \end{table*}
    
\section{Model characterization}
To further characterize the RaViTT strategy, we present example attention maps (Section \ref{sec-attentionmaps}), model efficiency ratios (Section \ref{sec-modelefficiency}), and validation loss function behavior (Section \ref{sec-trainingbehavior}).

\begin{figure}[ht!]
\begin{center}
\includegraphics[width=\linewidth]{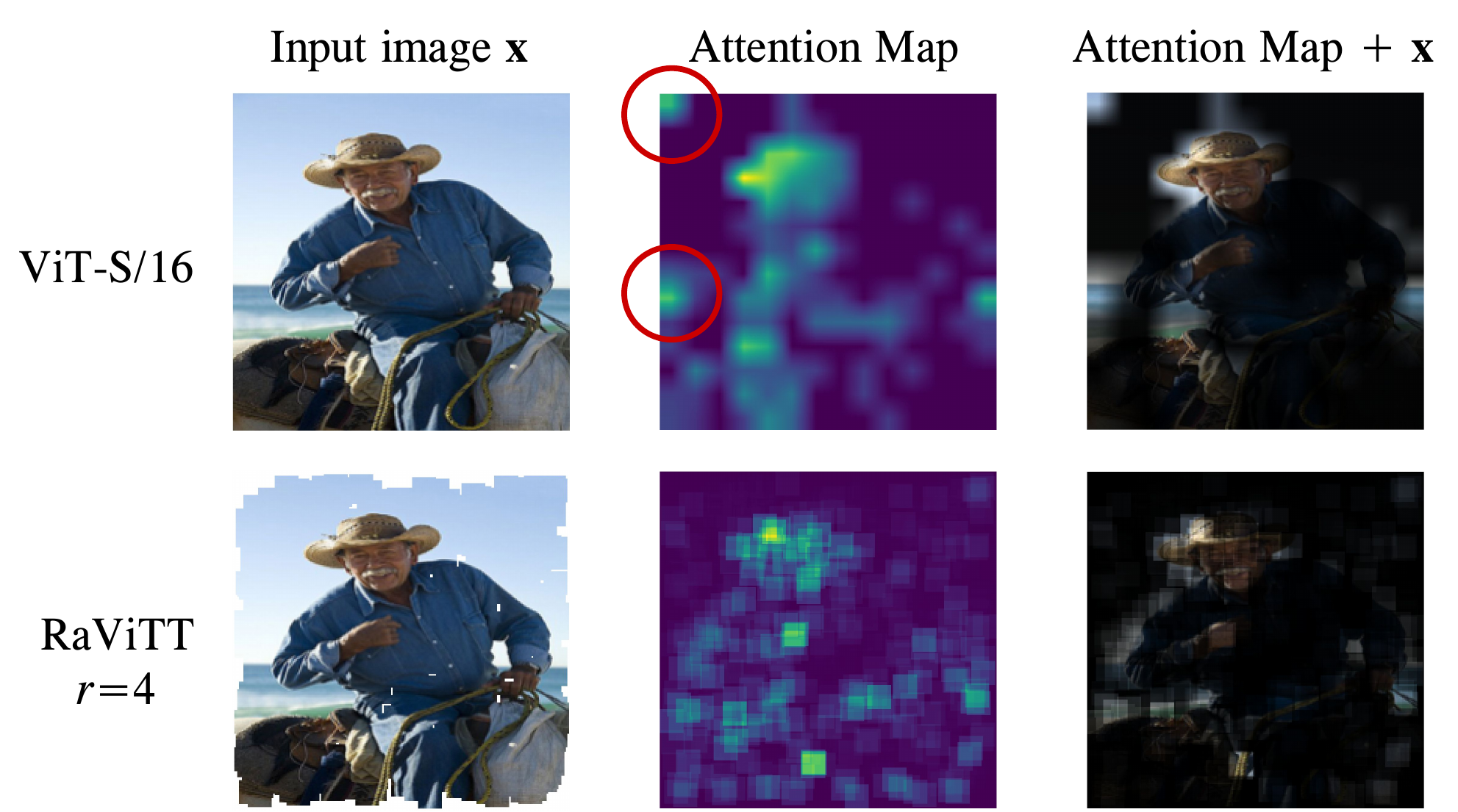}
\end{center} 
   \caption{Attention maps for an image of the validation split of ImageNet-1k (\textit{cowboy-hat} class). From left to right, input and results are shown for the baseline ViT-S/16 (top row) and RaViTT with $r=4$ (bottom row). Red circles highlight two regions of interest (attention spikes) that appear only with the baseline model.}
\label{fig:attn-maps}
\end{figure}
\subsection{Attention maps}
\label{sec-attentionmaps}
Attention maps are tools to visualize how ViTs capture relationships between input image patches, selectively focusing on relevant features while disregarding irrelevant ones. 
To generate attention maps, we used Attention Rollout \cite{abnar-zuidema-2020-quantifying}. The averaged attention was then normalized and transformed into a 2D heatmap.
The attention maps of the base ViT and RaViTT were rearranged into respective 2D images of $\sqrt{L}\times \sqrt{L}$ values re-scaled to $H\times W$ pixels. For RaViTT with \modea{}, the heatmap was constructed by mapping the patch attention levels to a color lookup table to  visualize accumulated attention.

In our assessing of the results where RaViTT and the baseline model differ and RaViTT yields the correct classification, we noticed that RaViTT tends to spread attention and produced spikes in different locations. Figure \ref{fig:attn-maps} illustrates this behavior by showing the attention maps of the baseline ViT and RaViTT \modea{}. Here, RaViTT did not produce attention spikes unrelated to the class (red circles). We also observed that, overall, RaViTT's attention appears more spread across heatmaps.

\begin{figure}[t]
\begin{center}
\includegraphics[width=\linewidth]{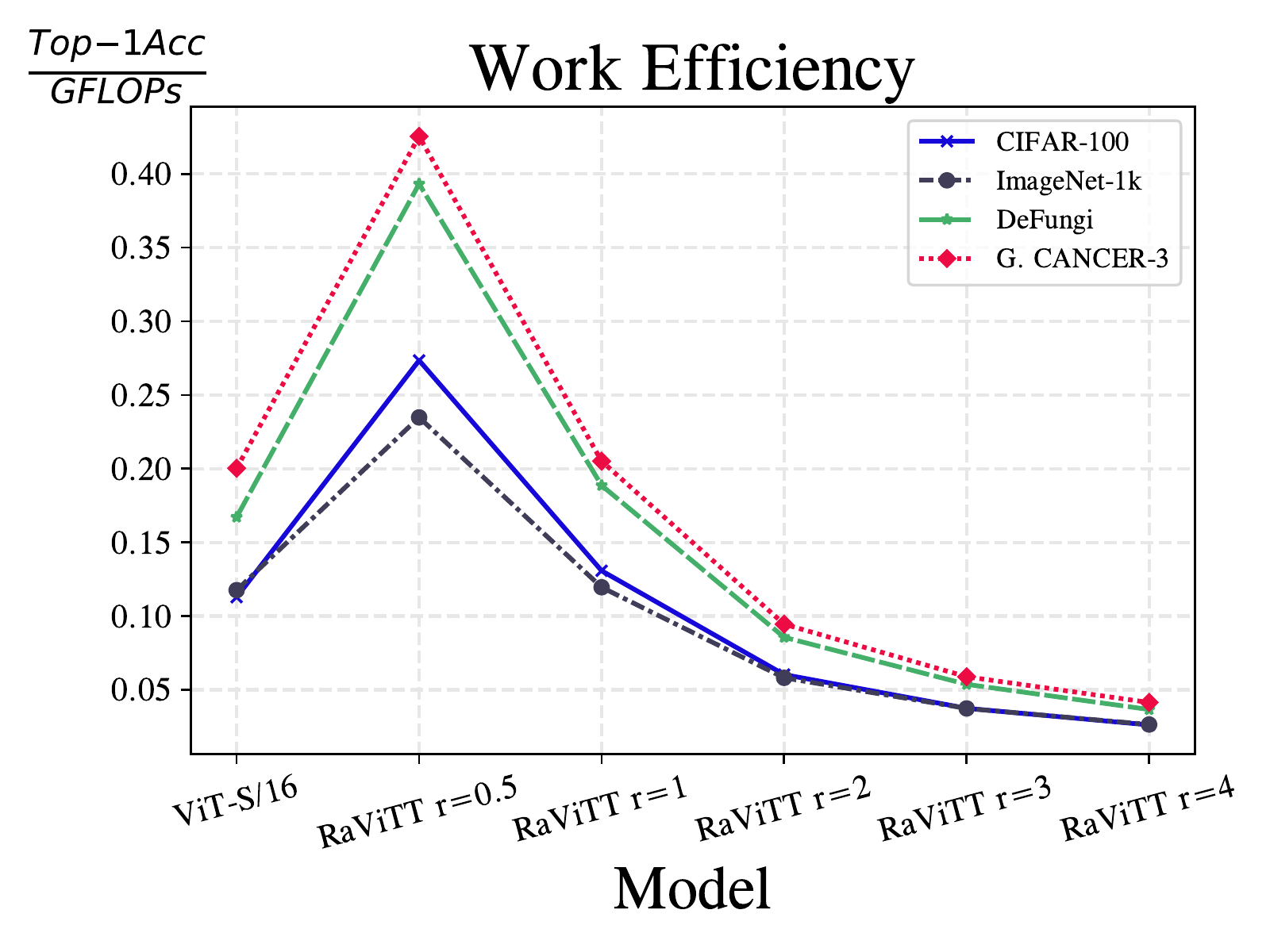}
\end{center}
   \caption{Baseline and RaViTT models' work efficiency across datasets. Higher is better.}
\label{fig:work-eff}
\end{figure}
\subsection{Work efficiency}
\label{sec-modelefficiency}
Given that in RaViTT the sampling factor $r$ can take any value as long as $r\cdot L \ge 0$, it is relevant to assess the impact of varying $r$ on the model's computational workload. Figure \ref{fig:work-eff} shows the Top-1 accuracy/GFLOPs ratio of each model across datasets. The results reveal that RaViTT with $r=1$, corresponding to the same number of tokens as the baseline ViT-S/16, achieves slightly higher efficiency in all datasets. Also, $r=0.5$ yields the highest efficiency in all datasets, which can be desirable in resource-limited scenarios. On the other hand, increasing $r$ results in a progressive decline in work efficiency.

\subsection{Validation loss function behavior}
\label{sec-trainingbehavior}
\begin{figure}[ht!]
\begin{center}
\includegraphics[width=\linewidth]{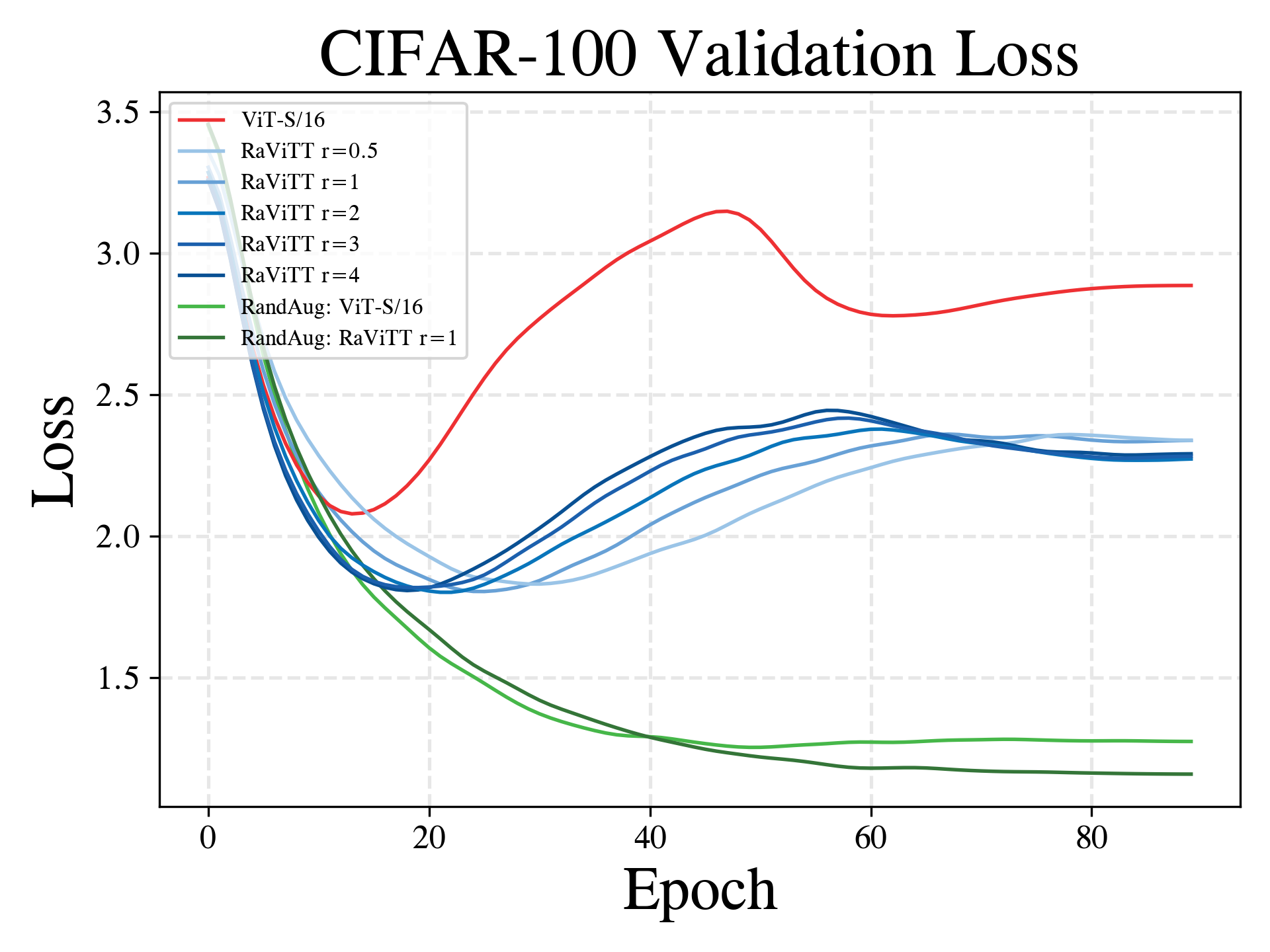}
\includegraphics[width=\linewidth]{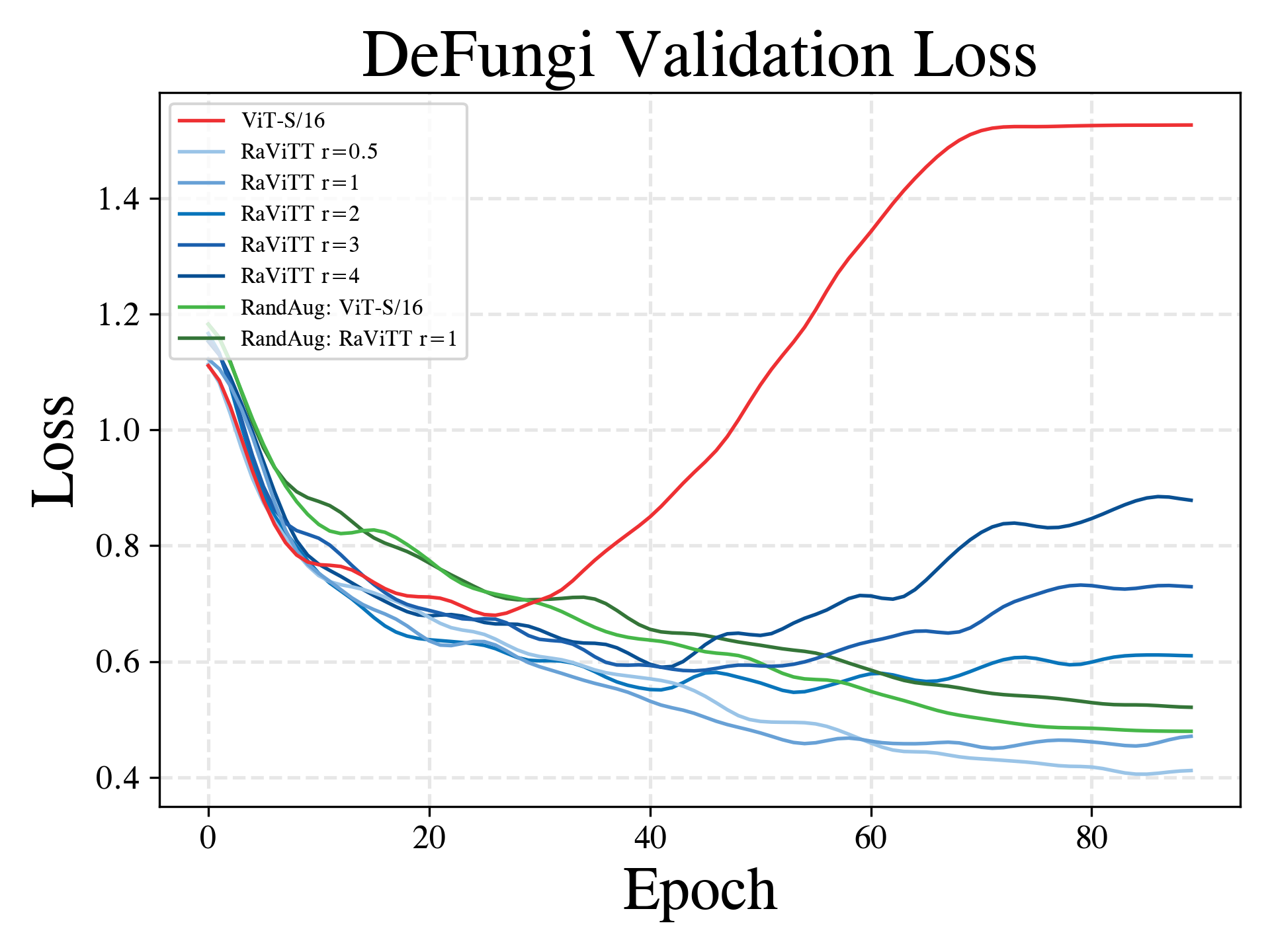}

\end{center}
   \caption{Validation loss for trained models. The top plot shows CIFAR-100 results, while the bottom shows results for the DeFungi dataset. RandAug refers to RandAugment. Lower is better.}
\label{fig:losses}
\end{figure}

Analyzing the training behavior of a model is crucial to assess its ability to generalize on new data and avoiding overfitting. Figure \ref{fig:losses} presents the validation loss curves for the CIFAR-100 (worst) and DeFungi (best) datasets. Starting with CIFAR-100, we observe 3 distinct overfitting behaviors: ViT-S/16 (red curve) shows the earliest overfit, while the RaViTT models (blue curves) exhibit overfit later on, and RandAugment (green curves) shows no sign of overfit during the fixed 90-epoch training. DeFungi plots also show three distinct curve patterns, but in this case, RaViTT (blue curves) for low $r$ values shows no overfit. Moreover, with $r=0.5$, it shows a behavior similar to ViT-S/16 RandAugment (light green curve). For the G. CANCER-3 and ImageNet-1k dataset, behaviors appear similar to the DeFungi dataset (not shown).


\section{Conclusions}
In all four datasets, RaViTT improves upon the baseline ViT-S/16 method. In the datasets with larger image sizes (ImageNet-1k, G. CANCER-3, DeFungi), this improvement is greater than the baseline ViT trained with SOTA data augmentation methods. Furthermore, RaViTT can improve accuracy with the same number of transformer tokens ($r=1$) or even less ($r<1$), meaning that higher accuracy can be achieved with less computational workload. This favorable feature of RaViTT may be a first step towards designing work-efficient ViT variants that optimize for  accuracy/GLOPs ratio.
Also, random distributions for patch sampling could be explored to improve the efficiency of RaViTT when the sampling factor increases ($r>1$).


\section*{Funding}
We acknowledge support of FONDECYT grants 1221696, 1221357, FONDEQUIP EQM210020, EQM180042.

{\small
\bibliographystyle{ieee_fullname}
\bibliography{egbib}

\begin{thebibliography}{10}\itemsep=-1pt

\bibitem{abnar-zuidema-2020-quantifying}
Samira Abnar and Willem Zuidema.
\newblock Quantifying attention flow in transformers.
\newblock In {\em Proceedings of the 58th Annual Meeting of the Association for
  Computational Linguistics}, pages 4190--4197, Online, July 2020. Association
  for Computational Linguistics.

\bibitem{ecancer}
Müller Bettina, García Carlos, Sola~José A, Fernandez Wanda, Werner Patrick,
  Cerda Mauricio, Slater Jeannie, Benavides Carlos, Arancibia Jorge, Ascui
  Rodrigo, Reyes Felipe, Stevens~Mary Anne, Miranda~Juan Pablo, Buchholtz
  Martin, and Corvalan Alejandro.
\newblock Perioperative chemotherapy in locally advanced gastric cancer in
  {C}hile: from evidence to daily practice.
\newblock {\em ecancer}, 15(1244), 2021.

\bibitem{vit_baseline}
Lucas Beyer, Xiaohua Zhai, and Alexander Kolesnikov.
\newblock Better plain {V}i{T} baselines for {I}mage{N}et-1k.
\newblock {\em arXiv preprint arXiv:2205.01580}, 2022.

\bibitem{Campanella2019}
Gabriele Campanella, Matthew~G. Hanna, Luke Geneslaw, Allen~P. Miraflor, Vitor
  Werneck~Krauss Silva, Klaus~J. Busam, Edi Brogi, Victor~E. Reuter, David~S.
  Klimstra, Thomas~J. Fuchs, and Thomas~J. Fuchs.
\newblock Clinical-grade computational pathology using weakly supervised deep
  learning on whole slide images.
\newblock {\em Nature Medicine}, 25(8):1301--1309, 2019.

\bibitem{chen2022scaling}
Richard~J Chen, Chengkuan Chen, Yicong Li, Tiffany~Y Chen, Andrew~D Trister,
  Rahul~G Krishnan, and Faisal Mahmood.
\newblock Scaling vision transformers to gigapixel images via hierarchical
  self-supervised learning.
\newblock In {\em Proceedings of the IEEE/CVF Conference on Computer Vision and
  Pattern Recognition}, pages 16144--16155, 2022.

\bibitem{randaugment}
Ekin~Dogus Cubuk, Barret Zoph, Jon Shlens, and Quoc Le.
\newblock Rand{A}ugment: Practical automated data augmentation with a reduced
  search space.
\newblock In H. Larochelle, M. Ranzato, R. Hadsell, M.F. Balcan, and H. Lin,
  editors, {\em Advances in Neural Information Processing Systems}, volume~33,
  pages 18613--18624. Curran Associates, Inc., 2020.

\bibitem{desir2012classification}
Chesner D{\'e}sir, Caroline Petitjean, Laurent Heutte, Mathieu Salaun, and Luc
  Thiberville.
\newblock Classification of endomicroscopic images of the lung based on random
  subwindows and extra-trees.
\newblock {\em IEEE transactions on biomedical engineering}, 59(9):2677--2683,
  2012.

\bibitem{dosovitskiy2020image}
Alexey Dosovitskiy, Lucas Beyer, Alexander Kolesnikov, Dirk Weissenborn,
  Xiaohua Zhai, Thomas Unterthiner, Mostafa Dehghani, Matthias Minderer, Georg
  Heigold, Sylvain Gelly, et~al.
\newblock An image is worth 16x16 words: Transformers for image recognition at
  scale.
\newblock {\em arXiv preprint arXiv:2010.11929}, 2020.

\bibitem{granziol2022learning}
Diego Granziol, Stefan Zohren, and Stephen Roberts.
\newblock Learning rates as a function of batch size: A random matrix theory
  approach to neural network training.
\newblock {\em J. Mach. Learn. Res}, 23:1--65, 2022.

\bibitem{Kanavati2020}
Fahdi Kanavati, Gouji Toyokawa, Seiya Momosaki, Michael Rambeau, Yuka Kozuma,
  Fumihiro Shoji, Koji Yamazaki, Sadanori Takeo, Osamu Iizuka, and Masayuki
  Tsuneki.
\newblock Weakly-supervised learning for lung carcinoma classification using
  deep learning.
\newblock {\em Scientific Reports}, 10, 06 2020.

\bibitem{Cifar_dataset}
Alex Krizhevsky, Geoffrey Hinton, et~al.
\newblock Learning multiple layers of features from tiny images.
\newblock {\em Technical Report.}, 2009.

\bibitem{liu2022tokenmix}
Jihao Liu, Boxiao Liu, Hang Zhou, Hongsheng Li, and Yu Liu.
\newblock Token{M}ix: Rethinking image mixing for data augmentation in vision
  transformers.
\newblock In {\em Computer Vision--ECCV 2022: 17th European Conference, Tel
  Aviv, Israel, October 23--27, 2022, Proceedings, Part XXVI}, pages 455--471.
  Springer, 2022.

\bibitem{liu2021iccv}
Ze Liu, Yutong Lin, Yue Cao, Han Hu, Yixuan Wei, Zheng Zhang, Stephen Lin, and
  Baining Guo.
\newblock Swin transformer: Hierarchical vision transformer using shifted
  windows.
\newblock In {\em Proceedings of the IEEE/CVF International Conference on
  Computer Vision (ICCV)}, pages 10012--10022, October 2021.

\bibitem{meng2022adavit}
Lingchen Meng, Hengduo Li, Bor-Chun Chen, Shiyi Lan, Zuxuan Wu, Yu-Gang Jiang,
  and Ser-Nam Lim.
\newblock Ada{V}i{T}: Adaptive vision transformers for efficient image
  recognition.
\newblock In {\em Proceedings of the IEEE/CVF Conference on Computer Vision and
  Pattern Recognition}, pages 12309--12318, 2022.

\bibitem{navarro2019color}
Carlos~F. Navarro and Claudio~A. Perez.
\newblock Color--texture pattern classification using global--local feature
  extraction, an {SVM} classifier, with bagging ensemble post-processing.
\newblock {\em Applied Sciences}, 9(15):3130, 2019.

\bibitem{SimpleDataAug}
Alexander~J Ratner, Henry Ehrenberg, Zeshan Hussain, Jared Dunnmon, and
  Christopher R{\'e}.
\newblock Learning to compose domain-specific transformations for data
  augmentation.
\newblock {\em Advances in neural information processing systems}, 30, 2017.

\bibitem{ren2022beyond}
Pengzhen Ren, Changlin Li, Guangrun Wang, Yun Xiao, Qing Du, Xiaodan Liang, and
  Xiaojun Chang.
\newblock Beyond fixation: Dynamic window visual transformer.
\newblock In {\em Proceedings of the IEEE/CVF Conference on Computer Vision and
  Pattern Recognition}, pages 11987--11997, 2022.

\bibitem{ImageNet_dataset}
Olga Russakovsky, Jia Deng, Hao Su, Jonathan Krause, Sanjeev Satheesh, Sean Ma,
  Zhiheng Huang, Andrej Karpathy, Aditya Khosla, Michael Bernstein,
  Alexander~C. Berg, and Li Fei-Fei.
\newblock {Image{N}et Large Scale Visual Recognition Challenge}.
\newblock {\em International Journal of Computer Vision (IJCV)},
  115(3):211--252, 2015.

\bibitem{Shorten2019}
Connor Shorten and Taghi~M. Khoshgoftaar.
\newblock A survey on image data augmentation for deep learning.
\newblock {\em Journal of Big Data}, 6(1):60, Jul 2019.

\bibitem{deit2021}
Hugo Touvron, Matthieu Cord, Matthijs Douze, Francisco Massa, Alexandre
  Sablayrolles, and Hervé Jégou.
\newblock Training data-efficient image transformers \& distillation through
  attention, 2020.

\bibitem{yun2019cutmix}
Sangdoo Yun, Dongyoon Han, Seong~Joon Oh, Sanghyuk Chun, Junsuk Choe, and
  Youngjoon Yoo.
\newblock Cut{M}ix: Regularization strategy to train strong classifiers with
  localizable features.
\newblock In {\em Proceedings of the IEEE/CVF international conference on
  computer vision}, pages 6023--6032, 2019.

\bibitem{zhang2017mixup}
Hongyi Zhang, Moustapha Cisse, Yann~N Dauphin, and David Lopez-Paz.
\newblock mixup: Beyond empirical risk minimization.
\newblock {\em arXiv preprint arXiv:1710.09412}, 2017.

\bibitem{DeFungi_dataset}
María Alejandra~Vanegas Álvarez, Leticia Sopó, Camilo Javier~Pineda Sopo,
  Farshid Hajati, and Soheil Gheisari.
\newblock P456 de{F}ungi: direct mycological examination of microscopic fungi
  images.
\newblock {\em Medical Mycology}, 60(Supplement\_1), 2022.

\end{thebibliography}
}

\end{document}